\def\year{2019}\relax
\documentclass[letterpaper]{article} %DO NOT CHANGE THIS
\usepackage{aaai19}  %Required
\usepackage{times}  %Required
\usepackage{helvet}  %Required
\usepackage{courier}  %Required
\usepackage{url}  %Required
\usepackage{graphicx}  %Required
\usepackage{multirow,paralist}
\usepackage{amssymb,mathtools}
\begin{document}
\title{xSense: Learning Sense-Separated Sparse Representations and Textual Definitions \\for Explainable Word Sense Networks}

\author{
  Ting-Yun Chang\thanks{The first three authors contribute equally to this work.}\quad Ta-Chung Chi\footnotemark[1]\quad Shang-Chi Tsai\footnotemark[1]\quad Yun-Nung Chen\\
Department of Computer Science and Information Engineering
  \\ National Taiwan University, Taipei, Taiwan \\
  \texttt{\{r06922168, r06922028, r06946004\}@ntu.edu.tw\quad y.v.chen@ieee.org}
}

%\author{
%  Ting-Yun Chang\thanks{The first three authors contribute equally to this work, the names are sorted alphabetically.}
%  \\ National Taiwan University \\
%  \texttt{r06922168@ntu.edu.tw}
%  \And
%  Ta-Chung Chi\footnotemark[1]
%  \\ National Taiwan University \\
%  \texttt{r06922028@ntu.edu.tw}
%  \And
%  Shang-Chi Tsai\footnotemark[1]
%  \\ National Taiwan University \\
%  \texttt{r06946004@ntu.edu.tw}
%  \And
%  Yun-Nung Chen
%  \\ National Taiwan University \\
%  \texttt{y.v.chen@ieee.org}
%}

%\author{AAAI Press\\
%Association for the Advancement of Artificial Intelligence\\
%2275 East Bayshore Road, Suite 160\\
%Palo Alto, California 94303\\
%}
\maketitle
\begin{abstract}
Despite the success achieved on various natural language processing tasks, word embeddings are difficult to interpret due to the dense vector representations.
This paper focuses on interpreting the embeddings for various aspects, including sense separation in the vector dimensions and definition generation.
Specifically, given a context together with a target word, our algorithm first projects the target word embedding to a high-dimensional sparse vector and picks the specific dimensions that can best explain the semantic meaning of the target word by the encoded contextual information, where the sense of the target word can be indirectly inferred.
Finally, our algorithm applies an RNN to generate the textual definition of the target word in the human readable form, which enables direct interpretation of the corresponding word embedding. 
This paper also introduces a large and high-quality context-definition dataset that consists of sense definitions together with multiple example sentences per 
polysemous word, which is a valuable resource for definition modeling~\cite{noraset2017definition} and word sense disambiguation.
The conducted experiments show the superior performance in BLEU score and the human evaluation test.
\end{abstract}

\section{Introduction}
Recently, machine learning models utilizing deep learning methodologies have achieved huge success on various tasks. However, state-of-the-art models are often extremely complex and have a huge amount of parameters such that transparency or interpretability are compromised.
Researchers cannot tell why or how the model makes a specific decision, which is particularly problematic when predictions are related to decision-critical applications such as medical applications.
Considering that understanding the underlying phenomenon in the model is critical,
\emph{interpretability}~\cite{lipton2016mythos} has therefore arisen as a key desideratum of machine learning models.

In natural language processing (NLP), word embeddings have produced significant improvement for different tasks.
However, the embeddings are dense representations that human finds difficult to interpret, which can be summarized in three main reasons:

\begin{compactenum}
\item \textbf{Polysemy}: Word embeddings mix different meanings into a single vector, which is also known as the polysemy issue~\cite{reisinger2010multi}.
\item \textbf{Dimension understanding}: The higher and lower values in the dimensions of an embedding vector are difficult to interpret and analyze for human~\cite{subramanian2017spine}.
%Human is often clueless as to what a ``high'' value along a given dimension of a vector signifies when compared to a “low” value~\cite{subramanian2017spine}.
\item \textbf{Semantic analysis}: We can only indirectly check the nearest neighbors to inspect the semantic meaning of a word embedding~\cite{noraset2017definition}.
\end{compactenum}

To address the polysemy issue, \citeauthor{arora2018linear} recently showed that a word embedding is the linear combination of its distinct sense embeddings weighted by the corresponding frequency in the training corpus~\cite{arora2018linear}. 
It proposed to use the weighted sum of multiple atoms of discourse to represent a word, where an atom indicates a concept.
Unfortunately, the discourse atom itself still suffers from the third issue and is not directly explainable.
Although it decomposed the vector representation into several atoms with their semantic meanings, it still suffered from the dimension understanding issue where the meaning of dimensions cannot be well explained.

In terms of the dimension understanding issue, several prior works attempted at projecting the dense embeddings into a sparse space and finding that words whose certain dimensions are large in a spare vector can form a semantic cluster~\cite{faruqui2015sparse,subramanian2017spine}.
Then they can isolate different senses into different dimensions and solved the first and the second issues together.
Nevertheless, inspecting nearest neighbors is still the only way to discover the meaning of a word embedding, so the semantic analysis issue remains unsolved.

Finally,~\citeauthor{noraset2017definition} tackled the semantic analysis issue by directly generating the textual definition of a word embedding based on a dictionary resource~\cite{noraset2017definition}. 
The main concern in this work is that they treated all word as monosemous and suffered from the polysemy issue.
\citeauthor{gadetsky2018conditional} tried to address this issue by training an encoder-decoder architecture along with a mask to generate context-dependent definitions.
However, both of these methods cannot explain the semantic meaning of the individual dimension (dimension understanding)~\cite{gadetsky2018conditional}.

\begin{table*}[t!]
\centering
\begin{tabular}{cll}
\hline
\bf Word & \multirow{1}{*}{\textbf{Definition}} & \textbf{Example Sentence} \\
\hline\hline
bass & The lowest adult male singing voice. & His \emph{bass} voice rings out attractively. \\
& & These are the opening words of the play, sung as a \emph{bass} solo.\\
\cline{2-3}
& The common European freshwater perch. & Only leisure anglers are allowed to fish \emph{bass} in Irish waters. \\
& & I did manage a couple of hours fishing a \emph{bass} pool the next morning.\\
\hline
\end{tabular}
%\vspace{-2mm}
\caption{Part of content for the word ``\emph{bass}'' in the proposed Oxford dataset.}
\label{table:oxford_example}
%\vspace{-1.8mm}
\end{table*}

Based on the above discussions, this paper proposes a novel explainable model, xSense, that embraces all benefits and avoids drawbacks.
That is, the proposed model can solve all three issues together.
The contributions of this paper are 4-fold:
\begin{compactitem}
\item Given a (context, word) pair, this paper can explicitly pin down the dimension in the sparse word representation that represents the sense of the word under a given context.
\item This paper is able to interpret the value of a specific dimension in the transformed sparse representation.
\item This paper provides the human understandable textual definition for a particular sense of a word embedding given its context.
\item We release a large and high-quality context-definition dataset that consists of abundant example sentences and the corresponding definitions for a variety of words.
\end{compactitem}

\section{Dictionary Corpus}
\label{sec:resource}
Dictionary corpora are usually available in the online electronic format.
However, they are often lack of example sentences. 
To the best of our knowledge, the Oxford online dictionary is the only one that contains an abundant amount of example sentences.\footnote{https://en.oxforddictionaries.com/} 
The prior work recently released a dataset based on this resource~\cite{gadetsky2018conditional}.
However, their dataset does not contain complete information achievable online, which hinders the usage for diverse tasks.
Some findings are described here:
1) Their dataset only provided one single example sentence for a definition, while there are usually multiple ones online.
2) Some provided example sentences in the dataset do not contain the target word, making the usage difficult.
3) Some provided example sentences do not align with their target word and the associated definitions.
Considering the quality of the released dataset, this paper addresses these problems from the prior work by releasing the newly-collected dataset and the toolkit for crawling the content.
A word example along with its multiple definitions and associated example sentences are shown in Table~\ref{table:oxford_example}.

\begin{table}[t!]
\centering
\begin{tabular}{lrr}
\hline
\bf Attribute & \bf Theirs & \bf Ours\\
\hline\hline
\#Words & 36,767 & 31,798 \\
Avg. \#sentence per def. & 1 & 27 \\
POS tag & N & Y\\
\hline
Total sentences & 122,319 & 1,299,821\\
Total tokens & 3,516,066 & 18,484,401\\
\hline
\end{tabular}
%\vspace{-2mm}
\caption{The dataset comparison between the prior work~\cite{gadetsky2018conditional} and the proposed one.}
\label{tb:dataset}
%\vspace{-1.8mm}
\end{table}

%while there are usually several example sentences provided along with a single definition, their dataset contains only a sentence. Second, some example sentences do not contain the target word at all, which is unreasonable. Third, some example sentences are totally uncorrelated to the target word and the corresponding definition, we even cannot find those sentences on the oxford dictionary website. To address these problems, we propose a new dataset and also release the toolkit we wrote to crawl the content with some examples demonstrated in table~\ref{table:oxford_example}. 
To be more specific, our dataset provides the following guarantees:
\begin{compactitem}
\item Each example sentence contains the target word it defines.
\item We include all example sentences of a specific definition available in the online dictionary.
\item We also include the corresponding POS tag of each word sense for further research usage.
\end{compactitem}
The statistics of the proposed dataset is summarized in Table~\ref{tb:dataset}, where it is obvious that our dataset contains much more example sentences, and the size is about 5 times larger than one provided by \citeauthor{gadetsky2018conditional}.
The high-quality and rich dataset can be leveraged in different NLP tasks, and this paper utilizes it for learning explainable word sense networks, xSense.
%in terms of total tokens.

\begin{figure*}[t!]
  \centering
  \includegraphics[width=0.86\linewidth]{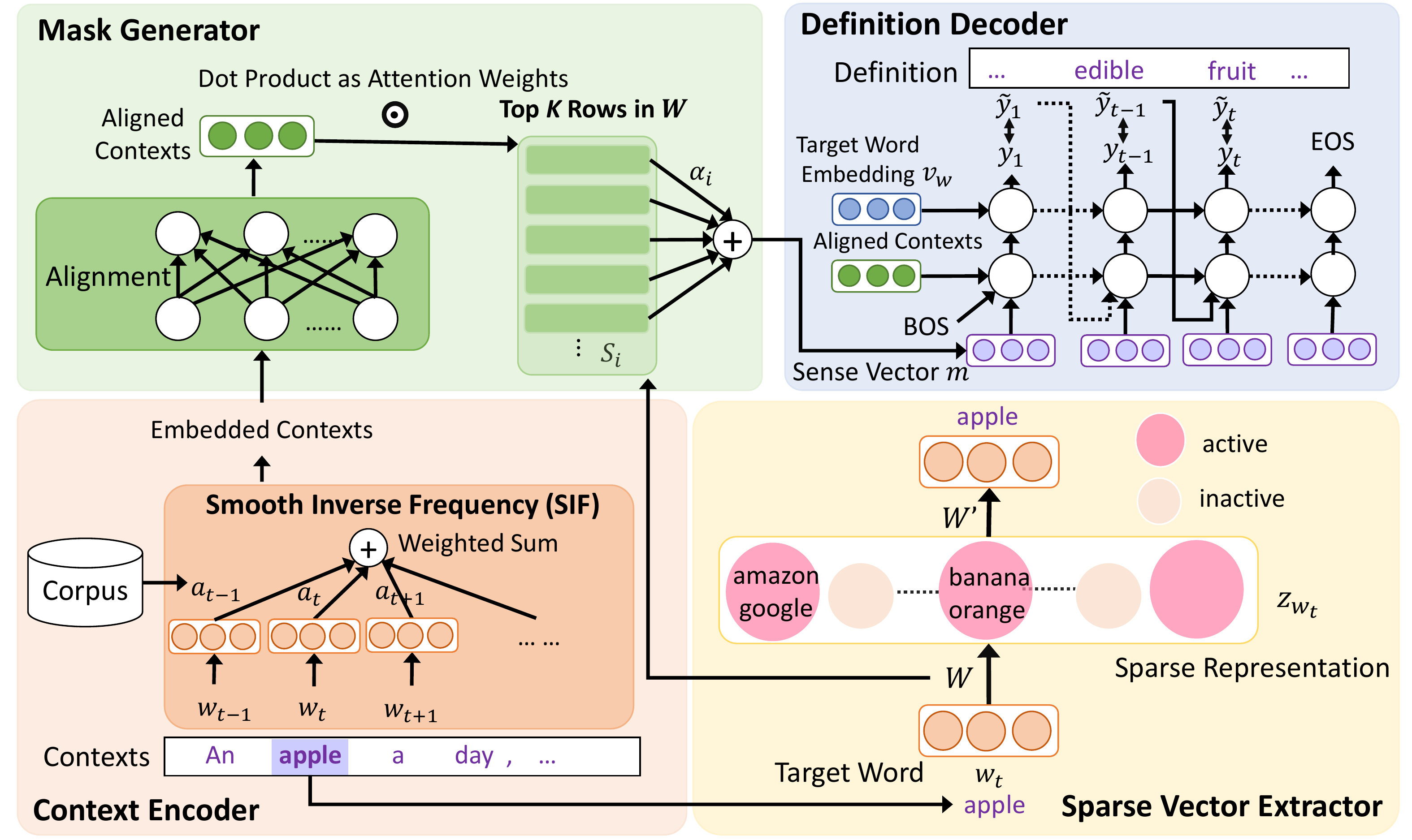}
  \caption{Illustration of the proposed xSense model. Encoder does not have parameters to train. The sparse extractor is pretrained and fixed during the training of mask generator and the decoder.}
  \label{fig:model}
\end{figure*}

\section{xSense: Explainable Word Sense Networks}
\label{sec:model}
The proposed model, xSense, consists of four main modules as illustrated in Figure~\ref{fig:model}.
Given a target word and its context, the model encodes the contexts (context encoder) and extracts its sparse representation (sparse vector extractor).
A mask is generated (mask generator) based on the contexts and the sparse vector in order to find the dimensions that encode the corresponding sense information, and then a definition sentence is generated (definition decoder).
Each component is detailed below.

\subsection{Dual Vocabularies}
We propose dual vocabularies, $V_{w2v}$ and $V_{dec}$, used in our model. 
The first one is the pretrained embeddings from word2vec\footnote{https://code.google.com/archive/p/word2vec/} used by the encoder and sparse vector extractor. 
The second one is randomly initialized and is only used by the decoder.
The goal of using two sets of vocabularies is to lower the out-of-vocabulary (OOV) rate. To be more specific, while $V_{w2v}$ contains lots of tokens, it misses some common functional words such as `a' and `of'.
In order to generate such common words in definition sentences, the dedicated vocabulary $V_{dec}$ is adopted.

\subsection{Context Encoder}
Given a context, the encoder module generates a distinguishable and meaningful sentence embedding. 
Because we do not assume additional resource for training the sentence embedding, the sentence is encoded in an unsupervised manner, which can be obtained in either sophisticated neural-based~\cite{kiros2015skip} or weighted-sum-based~\cite{arora2016simple} methods. 
The latter method is chosen in this paper due to two reasons.
First, neural-based methods require additional training data and much longer training time. 
Second, considering the goal of this paper is interpretability, weighted-sum method is more transparent for humans to interpret and investigate the error.\footnote{We also tried training a bidirectional GRU encoder, the performance is roughly the same.}

In our weighted-sum approach, we apply the smooth inverse frequency (SIF) embeddings~\cite{arora2016simple}, which is inspired by the discourse random walk model~\cite{arora2016latent}. 
Formally, given word embeddings ${{\bf v}_w:w\in V_{w2v}}$, a sentence $s\in S$, where $S$ is the set of all training sentences, a smoothing parameter $a$, and the occurrence probabilities ${p(w): w\in V_{w2v}}$ of the words derived from the training corpus, \emph{SIF} computes:
\begin{equation}
\label{eq:sif}
{\bf v}_s = \frac{1}{|s|}\sum_{w\in s}\frac{a}{a+p(w)}{\bf v}_w,
\end{equation}
where $|s|$ is the length of sentence $s$ and ${\bf v}_s$ will be used in the mask generator to generate the attention mask.

\subsection{Sparse Vector Extractor}
Words have large values in a specific dimension of their sparse representations often form a semantic cluster~\cite{faruqui2015sparse,subramanian2017spine}. 
This characteristic helps interpret the semantics in different dimensions.
Inspired by the idea about sparse coding in~\citeauthor{subramanian2017spine}, we incorporate a sparse vector extractor to learn the sparse representation of the target word~\cite{subramanian2017spine}:
\begin{align}
\label{eq:encoding_phase}
{\bf z}_w&=f({\bf W}_{enc}{\bf v}_w + {\bf b}_{enc})\\
\label{eq:decoding_phase}
{\bf v}'_w&={\bf W}_{dec}{\bf z}_w + {\bf b}_{dec}
\end{align}
where $f$ is the capped-ReLU activation function, ${\bf W}_{enc} \in {\rm I\!R}^{m\times d}$, ${\bf b}_{enc}\in {\rm I\!R}^{m}$, ${\bf W}_{dec} \in {\rm I\!R}^{d\times m}$, ${\bf b}_{dec}\in {\rm I\!R}^{d}$ are the learning parameters, and $d$ is the dimension of the word embedding.

This formulation follows a regular $k$-sparse autoencoder aiming at minimizing  \emph{reconstruction loss} and \emph{partial sparsity loss}~\cite{makhzani2013k,subramanian2017spine}.
\citeauthor{makhzani2013k} pointed out that the $k$-sparse autoencoder can be viewed as the variant of iterative thresholding with the inversion algorithm~\cite{maleki2009coherence}, which aims to train an overcomplete matrix $\bf W$ as orthogonal as possible. 
After training, $\bf W$ can be used as the~\emph{dictionary} in the sparse recovery stage.
In the context of word embeddings, the matrix ${\bf W}_{dec}$ contains the orthogonal basis of the embedding space, which are likely to be the basic semantic components.

We link this observation to the~\emph{discourse atom}, the basic sense component~\cite{arora2018linear}.
\citeauthor{arora2018linear} showed that a set of word embeddings can be disentangled into multiple discourse vectors by sparse coding.
Formally, given word embeddings ${{\bf v}_w}: w\in V_{w2v}$ in ${\rm I\!R}^d$ and an integer $m\gg d$, the goal is to solve:
\begin{equation}
\label{eq:discourse_obj}
\sum_{w\in V_{w2v}}\left\|{\bf v}_w-\sum_{j=1}^{m}\alpha_{w,j}{\bf A}_j\right\|_2^2,
\end{equation}
where $\alpha_{w,j}$ represents how much the discourse vector ${\bf A}_j$ weighs in constituting $v_w$. Both ${\bf W}_{dec}$ and the discourse atom are the basic semantic components of the embedding space. Moreover, from the viewpoint of matrix operation, (\ref{eq:discourse_obj}) is equivalent to (\ref{eq:decoding_phase}) with $\alpha_{w,j}={\bf z}_{w, j}$ and ${\bf A}_{j}={\bf W}_{dec,j}$, where ${\bf W}_{dec,j}$ is the $j$-th column of the matrix. 
In practice, since ${\bf z}_w$ is directly generated by ${\bf W}_{enc}$, we use the corresponding row vectors of ${\bf W}_{enc}$ in the mask generator.
As illustrated in Figure~\ref{fig:model}, the sparse vector extractor focuses on decomposing different senses into different dimensions via sparse coding, and the trained sparse encoder is for the mask generator usage.

\subsection{Mask Generator}
\label{sec:mask_gen}
The mask generator module is the key for interpretability, which connects the encoder and the sparse extractor and automatically finds the sense-specific dimensions. % while acting as the key for interpretability.
Given the~\emph{SIF} embedding ${\bf v}_s$ and a target word embedding ${\bf v}_w$, we focus on extracting the sense information from ${\bf v}_w$ according to its contexts.
${\bf v}_w$ is first fed into the sparse vector extractor to produce its sparse representation ${\bf z}_w$. 
We then lookup $K$ highest values in the sparse vector and retrieve the corresponding vectors in ${\bf W}_{enc}$, which is learned from the sparse vector extractor.
Formally, we compute the sparse representation of the target word by (\ref{eq:encoding_phase}) and obtain $K$ largest values:
\begin{equation}
\label{eq:argsort}
\gamma_{1\cdots K} = argsort_{K}({\bf z}_w).
\end{equation}
We retrieve the rows of ${\bf W}_{enc}$ according to the indices obtained in (\ref{eq:argsort}):
\begin{equation}
{\bf s}_j = {\bf W}_{enc}[\gamma_j], \quad j \in 1\cdots K.
\end{equation}
${\bf s}_j$ is therefore the $\gamma_j$-th row vector of ${\bf W}_{enc}$.
We calculate the inner product between the sentence embedding ${\bf v}_s$ and the basis vectors ${\bf s}_j$ to generate a weighted mask.
However, the direct calculation is unreasonable since they do not align well in the vector space.
Because both ${\bf v}_s$ and ${\bf s}_j$ are derived from the same pretrained embeddings by almost-linear operations, we assume that learning an additional linear transformation $\bf T \in {\rm I\!R}^{d \times d}$ can effectively align the space~\cite{conneau2017word}.
The inner product is thus calculated after the transformation:
\begin{equation}
\label{eq:transform}
d_{1\cdots k}={\bf T}{\bf v}_s \odot {\bf s}_j,\quad j\in 1\cdots K.
\end{equation}
The mask is calculated by a softmax layer:
\begin{equation}
\label{eq:att}
\alpha_j = \frac{\exp (d_j)}{\sum_j \exp (d_j)}, \quad j\in 1\cdots K.
\end{equation}
Finally, the retrieved basis vectors are weighted by the mask and then the sense vector is formed:
\begin{equation}
\label{eq:mh}
{\bf m} = \sum_j \alpha_j\cdot{\bf s}_j, \quad j\in 1\cdots K.
\end{equation}

\subsection{Definition Decoder}
The decoder module generates a textual definition for a target word given its context.
GRU is applied as our recurrent unit~\cite{cho2014learning}.
We denote a target definition sentence as a sequence of tokens: 
\begin{equation}
\label{eq:target_notation}
\tilde{\bf y} = \left\{\tilde{y_1}\text{,}\tilde{y_2}\text{,}\ldots\tilde{y_{M}}  \right\},
\end{equation}
where $M$ is the number of words in the definition.
We assign the aligned~\emph{SIF} embedding ${\bf Tv}_s$ to the initial hidden state of the first-layer GRU and the target word embedding ${\bf v}_w$ to the initial state of the second-layer GRU illustrated in Figure~\ref{fig:model}:
\begin{align}
\label{eq:gru_hidden}
{\bf h}^{1}_{0} =& {\bf Tv}_s, \\
\label{eq:gru_hidden2}
{\bf h}^{2}_{0} =& {\bf v}_w.
\end{align}
The goal of using the pretrained target word embedding as the initial hidden state is to provide explicit signal for the model in order to generate coherent and consistent definitions.
We also conduct the experiments using signals other than ${\bf v}_w$ in the experiment section~\ref{sec:exp} to analyze the effectiveness.
This initialization conditions the decoder to correctly generate definitions.
For each decoding step, the input to the cell is concatenated as:
\begin{equation}
\label{eq:gru_input}
{\bf x}_t = [{\bf v}_g,{\bf m}],
\end{equation}
where ${\bf v}_g \in V_{dec}$ is the ground truth word embedding at $t$-th timestep and {\bf m} is the sense vector calculated as (\ref{eq:mh}).
The decoding process terminates when an end-of-sentence token is predicted. The internal structure of a GRU cell is:
\begin{align}
\label{eq:decoding}
{\bf r}_t&=\sigma({\bf W}_{r}\cdot\left[{\bf h}_{t-1} , {\bf x}_{t}\right]),\\
{\bf z}_t&=\sigma({\bf W}_{z}\cdot\left[{\bf h}_{t-1} , {\bf x}_{t}\right]),\\
\tilde{{\bf h}_t}&=\tanh({\bf W}_{\tilde{h}}\cdot\left[{\bf r}_t*{\bf h}_{t-1}, {\bf x}_{t}  \right]),\\
{\bf h}_t&=(1-{\bf z}_t)*{\bf h}_{t-1}+{\bf z}_{t}*\tilde{{\bf h}_t}.
\end{align}
The output is generated by passing the hidden state through a linear layer:
\begin{equation}
{\bf O}_t={\bf W}_o\cdot{\bf h}_{t}.
\end{equation}
where ${\bf W}_o \in {\rm I\!R}^{|V_{dec}| \times d}$.
We use ${\bf O}_t$ to generate the final distribution over $V_{dec}$ via softmax operation. Formally,
\begin{align}
p_{t,i} =& \frac{{\exp({\bf O}_{t,i})}}{\sum_j \exp({\bf O}_{t,j})}, \\
y_t =& \arg\max_i p_{t,i}
\end{align}
Note that during the testing phase, the decoder is auto-regressive. 
Formally, (\ref{eq:gru_input}) becomes:
\begin{equation}
{\bf x}_t = [{\bf v}_{y_t},{\bf m}].
\end{equation}

\begin{table*}[t!]
  \centering
  \begin{tabular}{lccccccccccc}
    \hline
    \multicolumn{5}{c}{\multirow{2}{*}{\bf Methods}} & \multicolumn{3}{c}{\bf Datasets}\\
    \cline{6-8}
    & & & & & \bf Large & \bf Small & \bf Unseen\\
    \hline\hline
    \multicolumn{7}{l}{\it 1) Baseline w/o contexts}\\
    \multicolumn{4}{l}{Noraset et al.~\shortcite{noraset2017definition}} & & 33.8 / 36.3 & 30.5 / 32.7 & 12.0 / 13.3\\
    \hline
    \multicolumn{7}{l}{\it 2) Baseline w/ contexts}\\
    \multicolumn{4}{l}{Seq2Seq} & & 20.1 / 21.1 & 18.3 / 18.7 & 11.3 / 10.5\\
    \multicolumn{4}{l}{Gadetsky et al.~\shortcite{gadetsky2018conditional}} & & 26.0 / 31.6 & 25.5 / 30.4 & ~9.8 / 11.3\\
%     \hline
%     \multicolumn{2}{l}{\it 3) Proposed w/o transformation}\\
%     pretrain sparse extractor & 17.0 / 18.5 & 15.7 / 16.8 & 12.1 / 13.6\\
%     jointly train sparse extractor & 15.？ / 17.4 & 17.7 / 15.9 & N / N\\
    \hline
%    \multicolumn{6}{l}{\it 3) Proposed} \\
    % w/ transformation}\\
    \it 3) Proposed & & \it 1-Layer Init & \it 2-Layer Init & \it Each Time Input\\
    % mh/mapped-sif & 17.5 / 19.9 &  16.5 / 18.1 & {\bf 12.5} / 13.8\\
    \cline{2-5}
\multirow{5}{*}{xSense}  & SSS & Sense Vector & Sense Vector & Sense Vector & 14.8 / 17.0 & 14.4 / 15.9 & 12.1 / 13.3\\
     & AAS & Aligned Contexts & Aligned Contexts & Sense Vector & 20.6 / 23.0 & 18.6 / 20.3 & 12.4 / 13.9\\
  & TTS  & Target Word & Target Word & Sense Vector & 33.6 / 35.9 & 29.4 / 31.3 & 11.9 / 14.2\\
  % & SST  & Sense Vector & Sense Vector & Target Word & 34.5 / 36.4 & { 30.5} / 32.1 & 12.0 / 13.2\\
  \cline{2-8}
 & ATS & Aligned Contexts & Target Word & Sense Vector & { 37.2} / 39.7 & 30.1 / 32.0 & {\bf 12.7} / {\bf 14.5}\\ 
 & TAS & Target Word & Aligned Contexts & Sense Vector & {\bf 40.0} /{\bf 42.6} & {\bf 31.9} /{\bf 33.9}  & 12.4 /13.2\\
    \hline
  \end{tabular}
  \caption{BLEU and ROUGE-L scores for baselines and various proposed architectures. (BLEU / ROUGE-L:F1).}
  \label{table:exp}
\end{table*}

\iffalse
\begin{table*}[t!]
  \centering
  \begin{tabular}{lcccccccc}
    \hline
    \multirow{2}{*}{\bf Methods} & \multicolumn{3}{c}{\bf Datasets}\\
    \cline{2-4}
    & \bf Large & \bf Small & \bf Unseen\\
    \hline\hline
    \multicolumn{2}{l}{\it 1) Baseline w/o contexts}\\
    Noraset et al.~\shortcite{noraset2017definition} & 33.8 / 36.3 & {\bf 30.5} / {\bf 32.7} & 12.0 / 13.3\\
    \hline
    \multicolumn{2}{l}{\it 2) Baseline w/ contexts}\\
    Seq2Seq & 20.1 / 21.1 & 18.3 / 18.7 & 11.3 / 10.5\\
    Gadetsky et al.~\shortcite{gadetsky2018conditional} & 26.0 / 31.6 & 25.5 / 30.4 & ~9.8 / 11.3\\
%     \hline
%     \multicolumn{2}{l}{\it 3) Proposed w/o transformation}\\
%     pretrain sparse extractor & 17.0 / 18.5 & 15.7 / 16.8 & 12.1 / 13.6\\
%     jointly train sparse extractor & 15.？ / 17.4 & 17.7 / 15.9 & N / N\\
    \hline
    \multicolumn{2}{l}{\it 3) Proposed} \\% w/ transformation}\\
    % mh/mapped-sif & 17.5 / 19.9 &  16.5 / 18.1 & {\bf 12.5} / 13.8\\
    aligned SIF, aligned SIF / m & 20.6 / 23.0 & 18.6 / 20.3 & 12.4 / 13.9\\
    aligned SIF, target / m & {\bf 37.2} / {\bf 39.7} & 30.1 / 32.0 & {\bf 12.7} / {\bf 14.5}\\ 
    m, m / m & 14.8 / 17.0 & 14.4 / 15.9 & 12.1 / 13.3\\
    target, target / m & 33.6 / 35.9 & 29.4 / 31.3 & 11.9 / 14.2\\
    m, m / target & 34.5 / 36.4 & {\bf 30.5} / 32.1 & 12.0 / 13.2\\
    \hline
  \end{tabular}
  \caption{BLEU and ROUGE-L scores for baselines and various proposed architectures. (BLEU / ROUGE-L:F1)}
  \label{table:exp}
\end{table*}
\fi

\subsection{Optimization}
There are two losses for optimizing the sparse extractor, where the first loss is the reconstruction loss:
\begin{equation}
\mathcal{L}_R(D)=\frac{1}{|D|}\sum_{w\in D}|{\bf v}_w-{\bf v}_w'|^2,
\end{equation}
where $D$ is the size of the whole dataset, and the second loss is the partial sparsity loss~\cite{subramanian2017spine}:
\begin{equation}
\mathcal{L}_{PS}(D) = \frac{1}{|D|}\sum_{x\in D}\sum_{h}({\bf v}_{w,h}(1-{\bf v}_{w,h})).
\end{equation}
This loss encourages every dimension $h$ of ${\bf v}_w$ to be either 0 or 1.
Note that the sparse extractor module is pretrained and fixed.
In order to train the whole model in an end-to-end fashion, we minimize the negative log likelihood over maximum decoding steps $M$:
\begin{equation}
\mathcal{L}_{NLL} = -\sum_{t=1}^{M} \log p_{t}(\tilde{y_t}).
\end{equation}

\section{Experiments}
\label{sec:exp}

To evaluate our proposed model, we conduct various sets of experiments using our newly collected Oxford dataset.

\subsection{Setting}
\subsubsection{Hyperparameters}
Both $V_{w2v}$ and $V_{dec}$ have dimension 300. 
For the encoder, we fix the smoothing term $a$ in (\ref{eq:sif}) to $10^{-3}$ as recommended~\cite{arora2016simple}.
For the sparse vector extractor, the similar setup is adopted~\cite{subramanian2017spine}.
We choose $K=5$ in the mask generator.
The definition decoder is a two-layer GRU~\cite{cho2014learning} with the hidden size 300, where the optimizer used is SGD with the learning rate $0.1$ for training the sparse vector extractor and the mask generator. 
The Adam optimizer~\cite{kingma2014adam} with the default settings is applied to the decoder. 

\subsubsection{Testsets}
In the experiments, we want to demonstrate the ability of the proposed model in two difficulty levels.
\begin{compactitem}
\item \textbf{Easy}: The easier one is to test (seen words, unseen contexts).  Concretely, the \textbf{small} testset is the one proposed by~\citeauthor{gadetsky2018conditional} with 6,809 instances, while the~\textbf{large} testset is the one we collect with 42,589 instances. 
\item \textbf{Hard}: The harder one is to test (unseen words, unseen contexts) in the~\textbf{unseen} testset with 808 instances, which consists of all target words that are never seen during training.
\end{compactitem}

\subsubsection{Evaluation Metrics}
Two objective measures are reported, including BLEU~\cite{papineni2002bleu} up to 4-gram and F measure of ROUGE-L~\cite{lin2004rouge}. 
Considering that BLEU score has a lot of smoothing strategies, we decide to follow prior work~\cite{noraset2017definition,gadetsky2018conditional} and use the sentence-BLEU binary in the Moses library\footnote{http://www.statmt.org/moses/} for a fair comparison.
Both scores are averaged across all testing instances.

\subsubsection{Baselines}
Two sets of baseline approaches are compared, where the first one does not consider the contexts and the second one does.
The baseline without contexts is essentially a language model conditioning on the pretrained word embeddings $V_{w2v}$, which shares the same architecture in~\citeauthor{noraset2017definition}. 
We reimplement the model and train on our proposed dataset for fair comparison.
For baselines with contexts, we train the model proposed by~\citeauthor{gadetsky2018conditional} with their strongest settings on our dataset and a vanilla sequence-to-sequence model with both encoder and decoder being a two-layer GRU network.

\subsubsection{Proposed Variants}
We tried different input variants of (\ref{eq:gru_hidden}), (\ref{eq:gru_hidden2}),  and (\ref{eq:gru_input}) to see the effectiveness of inputting the explicit signal during decoding.
Specifically, for the 1-layer and 2-layer initialization of GRU and the additional input at each time step, different combinations of aligned contexts (A), the target word vector (T), and the sense vector (S) are attempted.
Note that at least one of the input should be the sense vector ($\bf m$ in (\ref{eq:mh})) in order to optimize the mask generator.
%The input of equation~\ref{eq:gru_hidden} can be either $\bf m$ or $\bf T \cdot {\bf v}_s$. The difference is that using ${\bf T}\cdot{\bf v}_s$ can give the decoder richer information about the context. The input of equation~\ref{eq:gru_input} should provide an explicit signal for the decoder to generate a consistent and coherent definition.

\begin{table*}[t!]
  \centering
  \begin{tabular}{|l|cc|cc|}
  \hline
\multirow{2}{*}{Model}  & \multicolumn{2}{c|}{\bf Top 1} & \multicolumn{2}{c|}{\bf Ranking Score}\\
  \cline{2-5}
   &  All & Multi-Sense & All & Multi-Sense\\
  \hline\hline
    Noraset et al.~\shortcite{noraset2017definition}  & 311 (30.8\%) & 17 (28.4\%) & 1887 (27.6\%) & 111(27.2\%) \\ 
  Gadetsky et al.~\shortcite{gadetsky2018conditional} & 240 (23.8\%) & 9 (15.0\%) & 1701 (24.9\%) & 92(22.5\%) \\
  xSense w/o Alignment &  115~~(11.4\%) & 8 (13.3\%) & 1182 (17.3\%) & 80 (19.9\%)\\
%  xSense-SST {\small (Sense Vector/Sense Vector/Target Word)} & 324 (24.3\%) & 17(22.1\%) & 1893 (21.7\%) & 101 (19.9\%)\\
  xSense-ATS {\small (Aligned Contexts/Target Word/Sense Vector)} &   \bf 342 (34.0\%) & \bf26(43.3\%) & \bf 2055 (30.2\%) & \bf 124 (30.4\%) \\
  \hline
  \end{tabular}
  \caption{Ranked human evaluation results on randomly sampled 200 questions from {\bf small} datasets.}
  \label{table:human_rank}
\end{table*}

\iffalse
\begin{table*}[t!]
  \centering
  \begin{tabular}{|l|cccc|cccc|}
  \hline
\multirow{2}{*}{Model}  & \multicolumn{4}{c|}{\bf Top 1} & \multicolumn{4}{c|}{\bf Ranking Score}\\
  \cline{2-9}
   & \#1 & \#2 & \#3 & Total & \#1 & \#2 & \#3 & Total\\
  \hline\hline
    Noraset et al.~\shortcite{noraset2017definition} & 111 & 104 & ~96 & 311 (23.3\%) & 612 & 639 & 636 & 1887 (21.6\%)\\ 
  Gadetsky et al.~\shortcite{gadetsky2018conditional} & ~77 & ~83 & ~80 & 240 (18.0\%) & 582 & 572 & 547 & 1701 (19.5\%)\\
  xSense w/o Alignment & ~40 & ~36 & ~39 & 115~~(8.6\%) & 401 & 384 & 397 & 1182 (13.6\%)\\
  xSense-SST {\small (Sense Vector/Sense Vector/Target Word)} & 117 & 104 & 103 & 324 (24.3\%) & 626 & 622 & 645 & 1893 (21.7\%)\\
  xSense-ATS {\small (Aligned Contexts/Target Word/Sense Vector)} & 140 & 100 & 102 & \bf 342 (25.7\%) & 803 & 617 & 635 & \bf 2055 (23.6\%)\\
  \hline
  \end{tabular}
  \caption{Ranked human evaluation results on randomly sampled 200 questions from {\bf small} datasets.}
  \label{table:human_rank}
\end{table*}
\fi

\subsection{Results}
The results are shown in Table~\ref{table:exp}.
Among all baselines, \citeauthor{noraset2017definition}'s work is the strongest baseline even though their model generates exactly the same definition regardless of different contexts.
The probable reason is that dictionary definitions are often written in a highly structured and similar format, thus generating the same definition for all contexts can still share some common words with the ground truth. % not clear here, maybe an example?

Among baselines leveraging contexts, the performance of the sequence-to-sequence model is worse than~\citeauthor{gadetsky2018conditional}'s.
The probable reason is that \citeauthor{gadetsky2018conditional} introduced a mask to differentiate different contexts and generate definitions accordingly. 
However, their performance is the worst among all models on~\textbf{unseen} dataset, which explicitly evaluates the generalizability.
The observation tells that the better performance on~\textbf{large} and~\textbf{small} are likely because of memorizing the information from the training data (overfitting).
In addition, the performance gain compared with \citeauthor{noraset2017definition}'s work reported in~\cite{gadetsky2018conditional} is only 0.46 of BLEU (full 100), which is insignificant.

To analyze the information richness of different variants, for two hidden layers, we replace the sense vector as initialization with the aligned contexts.
%For our proposed xSense, we can see that using the aligned contexts (${\bf Tv_s}$) as the initial hidden state of the 1-layer in the decoder outperforms the one using the sense vector (${\bf m}$) (ATS v.s. TTS in Table~\ref{table:exp}).
Comparing between SSS and AAS in Table~\ref{table:exp}, using the aligned contexts (${\bf Tv_s}$) as the initial hidden state in the decoder outperforms the one only using the sense vector (${\bf m}$).
The reason is that the aligned contexts provide the decoder additional information of contexts and help generate more sophisticated definitions, while the sense vector is the weighted sum of basis vectors as shown in (\ref{eq:mh}), which may introduce some errors due to the imperfectness of the sparse vector extractor.

We also try to replace the sense vector with pretrained target word embedding to initialize the hidden state of the decoder, and the significantly better performance is observed (SSS v.s. TTS). 
This is reasonable because pretrained embeddings are trained on a large corpus and thus contain robust and rich information.
In addition, it provides a static representation that stabilizes the training process of the decoder.
However, we find out that while having good performance on BLEU/ROUGE scores, the variety of generated definitions is lower than the one of the aligned contexts. 
In other words, despite pretrained word embeddings being informative, its semantic meaning is likely dominated by the most frequent senses in the training corpus. 
In fact, we observe that simply using the target word embedding as the initial decoder hidden state cannot distinguish the difference between fine-grained senses; The definitions generated by TTS are the major senses in most testing instances.

Finally, to balance between variety and correctness, combining aligned contexts with pretrained word embedding as our decoder initialization (ATS, TAS) is a natural choice from the experiments.
The result is the best one for \textbf{Large} and \textbf{Unseen} datasets, demonstrating better performance and generalizability.

The performance is poorer for all models on \textbf{Unseen} rather than other testsets. That is because these words are not encountered during training, making the embedding explanation much more difficult.
Moreover, we manually check the test words and find out that most of them are uncommon words, making this testset even harder.

\subsection{Human Evaluation}
In order to justify the quality of the generated definitions, we randomly select two hundred samples from the \textbf{Small}  dataset for human evaluation, where two settings are reported, one includes all words (All) and another includes only the words whose multiple($\ge$3) senses are sampled (Multi-Sense).
There are four candidate models including all baselines, one of our best models (xSense-ATS), and xSense without alignment in (\ref{eq:transform}) that jointly learns the sparse vector extractor.
Three human annotators are recruited to rank the generated definitions given the target word and its corresponding contexts in each sample. 
Table~\ref{table:human_rank} shows the final statistics, where the top 1 choice and the accumulated scores are reported (4: first, 3: second, 2: third, 1: last).
%, where the agreement is about xxx.
Note that in some samples, two models may generate exactly the same definition and if an annotator picks either of them, we assign the same score to another.

It can be found that our model performs best among all candidates for both settings of all target words and multi-sense target words.
While \citeauthor{subramanian2017spine}'s work achieves the second-best performance, it cannot distinguish different senses since it does not consider the contexts, which makes the task about explainable embeddings entirely useless. The multi-sense setting indeed shows that our proposed model significantly outperforms theirs.
The worst model is the one without alignment, indicating that the basis vectors and the sentence embedding do not align in the vector space so that the attention cannot be correctly obtained.

\iffalse
\begin{table*}[t!]
  \centering
  \begin{tabular}{lcccc}
  \hline
  \bf Annotator & w/o transform & \cite{noraset2017definition} & \cite{gadetsky2018conditional} & m/ target \\
  \hline
  \#1 & 40 & 111 & 77 & 117 \\
  \hline
  \#2 & 36 & 104 & 83 & 104 \\
  \hline
  \#3 & 39 & 96 & 80 & 103 \\
  \hline
  Total &  115 & 311 & 240 & 324\\
  \hline
  \end{tabular}
  \caption{Human evaluation results on randomly sampled 200 questions from {\bf small} datasets}
  \label{table:human}
\end{table*}

\begin{table*}[t!]
  \centering
  \begin{tabular}{lcccc}
  \hline
  \bf Annotator & w/o transform & \cite{noraset2017definition} & \cite{gadetsky2018conditional} & m/ target \\
  \hline
  \#1 & 401 & 612 & 582 & 626 \\
  \hline
  \#2 & 384 & 639 & 572 & 622 \\
  \hline
  \#3 & 397 & 636 & 547 & 645 \\
  \hline
  Total &  1182 & 1887 & 1701 & 1893 \\
  \hline
  \end{tabular}
  \caption{Ranked human evaluation results on randomly sampled 200 questions from {\bf small} datasets}
  \label{table:human_rank}
\end{table*}
\fi

\begin{table*}[t!]
\centering
\begin{tabular}{|c|l|c|c|}
\hline
\multirow{1}{*}{\textbf{Target Word}} & \bf Contexts, Generated Definition, Nearest Neighbors \\
\hline\hline
\multirow{6}{*}{band} & He looked around and saw what he was looking for a \emph{band} of thin electrical wire.\\
& {\bf Gen. Definition}: A circular revolving plate supporting a single wire or other object of rock \\
& {\bf Nearest Neighbors}: inductor, chipset, transceiver (701-th dimension)\\
\cline{2-2}
&  In her spare time she performs as one of three vocalists in a \emph{band}.\\
& {\bf Gen. Definition}: A group of musicians actors or dancers who perform together\\
& {\bf Nearest Neighbors}: punk, tracklist, hiphop (215-th dimension)\\
\hline
\multirow{6}{*}{cool} & I closed my eyes again and imagined myself in a \emph{cool} refreshing blue pool.\\
 & {\bf Gen. Definition}: soothing or refreshing because of its low temperature\\
 & {\bf Nearest Neighbors}: humid, moist, wintry (213-th dimension)\\ 
\cline{2-2}
&  There is need to \emph{cool} off our tempers and stop fanning the embers of dissent.\\
& {\bf Gen. Definition}: unemotional undemonstrative or impassive dancers who perform together \\
& {\bf Nearest Neighbors}: levelheaded, gentlemanly, personable (161-th dimension)\\ 
\hline
\multirow{6}{*}{bow} & It was customary when they finished  to \emph{bow} as a sign of respect to their master.\\
 & {\bf Gen. Definition}: a gesture of acknowledgement or concession to \\
 & {\bf Nearest Neighbors}: palanquin, casket, limousine (143-th dimension)\\
 \cline{2-2}
&  Pat was wearing a black spandex long sleeved shirt with a thin thread tied in a \emph{bow}\\
& {\bf Gen. Definition}: a length of cord rope wire or other material serving a particular purpose  \\
& {\bf Nearest Neighbors}: embroidery, ribbon, fabric (782-th dimension)\\ 
\hline
\end{tabular}
%\vspace{-2mm}
\caption{The analysis of the generated definition and the highest value of a single dimension in a sparse vector.}
\label{table:attention}
\end{table*}

\subsection{Qualitative Analysis}

%\subsubsection{Attention Visualization}
An important capability of our model is that we can pin down the dimension in the sparse representation of a target word given its context. 
This is difficult to tell in numbers, so we show some samples for analysis in Table \ref{table:attention}. 
We can see that the nearest neighbors and the generated definitions belong to the same semantic clusters.
Moreover, we are able to disentangle multiple senses based on the given contexts.

\begin{table*}[t!]
\centering
\begin{tabular}{|c|l|c|c|}
\hline
\multirow{1}{*}{\textbf{Target Word}} & \bf Contexts, Ground Truth, Generated Definition, Nearest Neighbors \\
\hline\hline
\multirow{4}{*}{bass} & Don't worry if all your \emph{bass} have been what we call schoolie bass which are fish under two or three pounds.
\\
& {\bf Ground Truth}: The common European freshwater perch \\
& {\bf Gen. Definition}: A bass guitar or double bass. (X)\\
& {\bf Nearest Neighbors}: yacht, vessel, surf, sail (148-th dimension) \\
\hline
\multirow{4}{*}{tie} & I of course immediately asked him how many knots he could \emph{tie}.
\\
& {\bf Ground Truth}: form a knot or bow in a ribbon lace \\
& {\bf Gen. Definition}: form a knot or bow in a ribbon lace \\
& {\bf Nearest Neighbors}: unbeaten, tiebreaker, victor (780-th dimension) (X) \\
\hline
\end{tabular}
\caption{Error analysis of common mistakes made by xSense.}
\label{table:error}
\end{table*}

To better understand the limitations of our model, we show some common mistakes in Table~\ref{table:error}. 
For the word~\emph{bass}, our model generates the wrong definition while picking up the correct nearest neighbors. 
Note that the generated wrong definition is another sense of~\emph{bass}, so the cause of this error may be due to the imbalance of sense frequency in training data, considering that \emph{bass} as a kind of fish is a relatively rare sense. 
For the word~\emph{tie}, the generated definition is correct while the selected nearest neighbors are wrong.
Because the nearest neighbors are determined by (\ref{eq:att}), this error type may be propagated from the \emph{SIF} sentence embedding.

\section{Related Work}
This work can be viewed as a bridge that connects sparse embeddings and sense embeddings together for better interpretability via definition modeling. 

\vspace{-1mm}
\paragraph{Sparse embedding}
Several works have shown that introducing sparsity in word embedding dimensions improves dimension interpretability
\cite{nonnegative,fyshe2015compositional} and the benefit of word embeddings as features in downstream tasks~\cite{guo2014revisiting}. 
They focused on investigating the internal characteristics of word embeddings, making it hard to perform real-world applications such as word sense disambiguation (WSD).
In addition, they cannot provide explicit textual definitions of word embeddings.

\vspace{-1mm}
\paragraph{Sense-level embedding}
In literature, most of the prior works assigned a vector representation for each sense of a word.
The work often assumed a large training corpus to facilitate the training of multi-sense embeddings in an unsupervised manner~\cite{reisinger2010multi,li2015multi,lee2017muse}.
Note that the sense embeddings in our framework are disentangled internally by a sparse autoencoder.
In this paper, the additional training data is not required.
Also, unlike the prior work, our model can provide human-readable definitions for better interpretability.

\vspace{-1mm}
\paragraph{Dictionary definition task}
There are several works that utilized dictionary definitions to perform the ranking task or learn word embeddings.
In the ranking tasks, the models are evaluated by how well
they rank words for given definitions~\cite{hill2015learning} or definitions for words~\cite{noraset2017definition}. 
Aside from ranking tasks,~\citeauthor{bahdanau2017learning} suggested using definitions to compute embeddings for out-of-vocabulary words.
Different from them, this paper focuses on utilizing the textural definitions to provide the capability of explaining the embeddings via human understandable natural language.

\section{Conclusion}
\label{sec:conclusion}
In this paper, the interpretability of word embedding dimensions is investigated. 
Our proposed model is able to pin down a specific dimension on its sparse representation via an attention mechanism in an unsupervised manner and generate its corresponding textual definition at the same time. 
In the experiments, the proposed model outperforms others for both quantitative results and human evaluation.
Finally, we release a new and high-quality dataset which is five times larger than the currently available one, providing potential directions for future research work.

\newpage
\bibliography{aaai}
\bibliographystyle{aaai}

\end{document}